\documentclass{bmvc2k}

\usepackage{times}
\usepackage{epsfig}
\usepackage{graphicx}
\usepackage{algorithm}
\usepackage[noend]{algpseudocode}
\usepackage{amsmath}
\usepackage{amssymb}
\usepackage{verbatim}
\usepackage{color}
\usepackage{multirow}
\usepackage{pifont}% http://ctan.org/pkg/pifont

\title{A Novel Incremental Cross-Modal Hashing Approach}

\addauthor{Devraj Mandal}{devrajm@iisc.ac.in}{1}
\addauthor{Soma Biswas}{somabiswas@iisc.ac.in}{1}

\addinstitution{
 Department of Electrical Engineering\\
 Indian Institute of Science\\
 Bangalore, India-560012.
}

\runninghead{Devraj}{A Novel Incremental Cross-Modal Hashing Approach}

\begin{document}

\maketitle

% *************************************************************************************************
%%%%%%%%% ABSTRACT
\begin{abstract}
	%Cross-modal retrieval is an important problem under study for the computer vision community.
	Cross-modal retrieval deals with retrieving relevant items from one modality, when provided with a search query from another modality. 
	Hashing techniques, where the data is represented as binary bits have specifically gained importance due to the ease of storage, fast computations and high accuracy.
	In real world, the number of data categories is continuously increasing, which requires algorithms capable of handling this dynamic scenario.
	In this work, we propose a novel incremental cross-modal hashing algorithm termed "iCMH", which can adapt itself to handle incoming data of new categories.
	The proposed approach consists of two sequential stages, namely, learning the hash codes and training the hash functions.
	At every stage, a small amount of old category data termed ``exemplars" is is used so as not to forget the old data while trying to learn for the new incoming data, i.e. to avoid catastrophic forgetting.
	In the first stage, the hash codes for the exemplars is used, and simultaneously, hash codes for the new data is computed such that it maintains the semantic relations with the existing data.
	For the second stage, we propose both a non-deep and deep architectures to learn the hash functions effectively.
	Extensive experiments across a variety of cross-modal datasets and comparisons with state-of-the-art cross-modal algorithms shows the usefulness of our approach.
\end{abstract}
% *************************************************************************************************

%%%%%%%%% BODY TEXT
\section{Introduction}

Cross-modal retrieval has become increasingly important due to the easy availability of multimedia data like images, textual documents, audio snippets, video files, etc.
The objective is to retrieve a related item of a particular modality (eg. video file),  given a query from some other modality (eg. audio snippet) and different approaches have been proposed for this problem.
One standard approach is to project data from the different modalities into a common latent space to enable comparisons between them \cite{ccca} \cite{gma} \cite{acmr} \cite{gssl} \cite{sscmlp}.
Different approaches have been proposed for unsupervised, semi-supervised as well as the supervised paradigm.
Hashing approaches, on the other hand hand represent the data from the different modalities using a binary representation such that the associations between the different data categories are preserved. \cite{seph} \cite{gsph} \cite{growbit}.

Most of these approaches assume that the data from all the categories are available during training.
This assumption may be restrictive in real-world, where data from new categories arise dynamically.
Retraining the cross-modal algorithm from scratch every time new category data is encountered is very inefficient.
Simple fine-tuning of the pre-trained model on the new category data may lead to catastrophic forgetting~\cite{lwf} \cite{icarl}, which results in severe degradation in retrieval performance for the old categories.
Thus it is imperative to build retrieval models which can be updated to handle incoming streams of new category data, while retaining the information from the existing categories.
Recently, for image classification problem, this has received significant attention~\cite{lwf} \cite{icarl}\cite{rebalance_cvpr2019}\cite{incremental_cvpr2019}\cite{lsil_cvpr2019}, but it is relatively unexplored for the cross-modal problem~\cite{lifelong}

In this work, we address the problem of incremental cross-modal retrieval, where data from new categories can come sequentially, and need not be present during the initial training.
The proposed framework iCMH (Incremental Cross-Modal Hashing) has two stages, namely (i) hash code learning and  (ii) hash function learning.
For the first stage, we propose an adaptive hash model which can be updated to efficiently learn the hash codes of the incoming data from the new categories, while using the hash codes of a small portion of the old category data ({\em exemplars}).
For the second stage, we propose both a non-deep and deep based solution and investigate distillation loss, class weighted loss and imbalanced data sampling strategies to overcome catastrophic forgetting.
The proposed framework works well under both the cross-modal retrieval and hashing paradigms.

The incremental learning problem can be broadly classified into two different categories - multi-class and multi-task \cite{lwf} \cite{rebalance_cvpr2019}.
In the \textit{multi-task setting}, the model is trained to work on a variety of different objectives with each objective relating to a specific set of categories \cite{lwf}.
Here, during testing the evaluation is done per task \cite{lwf} \cite{lifelong} with respect to the data of relevant set of categories.
The \textit{multi-class setting} \cite{lwf} \cite{rebalance_cvpr2019} \cite{icarl} is a more realistic and challenging scenario where the evaluation is done over all the classes seen till date.
In this work, we focus on the more challenging \textit{multi-class setting} for cross-modal retrieval and hashing applications.
The main contributions of this work are as follows.
\begin{enumerate}
	\item We investigate how standard cross-modal algorithms perform under the multi-class retrieval setting when data spread over new categories are available in an incremental fashion.	
	\item We propose a novel incremental cross-modal hashing algorithm, whose model parameters can be updated to reflect new incoming data without suffering from catastrophic forgetting.
	\item We propose a non-deep and a deep based methodology to solve the above objective.
	\item We investigate ways to counter the phenomenon of catastrophic forgetting by using different distillation losses, class weighted classification losses and imbalanced data sampling strategies.
	\item Extensive experiments on four standard cross-modal datasets shows the effectiveness of the proposed framework.
\end{enumerate}

%Various algorithms have been developed for cross-modal retrieval applications that work well under the unsupervised, semi-supervised and supervised paradigm.
%In general, supervised algorithms tend to outperform their unsupervised counterparts due to the presence of the label annotation.
%The semi-supervised algorithms treads a middle ground by trying to leverage the information from the vast amount of unlabeled data and gives better performance than the unsupervised methods while requiring considerably less amount of labeled data.
%Various methods \cite{icarl} \cite{rebalance_cvpr2019} \cite{incremental_cvpr2019} \cite{lsil_cvpr2019} have been developed to deal with ``catastrophic forgetting" like preserving - (1) parameters of the old model, (2) knowledge of the old categories by using specific losses like distillation loss, (3) and handling imbalance between the old class examples and new category data.
%To this end, generally, a small fraction of the old category data called ``exemplars" is suitably selected and used to transfer the knowledge from the old model to the new model by the procedure of knowledge distillation \cite{icarl} \cite{rebalance_cvpr2019} \cite{incremental_cvpr2019} \cite{lsil_cvpr2019}.

% ***********************************************************************************

\section{Related Work}

Cross-modal retrieval \cite{ccca} \cite{gssl} \cite{sscmlp} \cite{gma} \cite{acmr} and hashing \cite{gsph} \cite{seph} \cite{growbit} have received significant attention because of numerous practical applications.
%The basic objective of each of the above method is to project the different domains data into the common domain where comparisons can be carried out.
%Hashing based approaches \cite{gsph} \cite{seph} \cite{growbit} design the common domain as a binary representation for efficient and fast retrieval.
%Methods have been developed in the unsupervised, semi-supervised and supervised paradigm.
In this work, we focus on the supervised paradigm, where the model needs to be updated to handle incoming data from the new categories.

\textbf{Supervised} methods~\cite{acmr} \cite{ccca} \cite{gma} uses the labels to learn the discriminative common domain representation and generally give better performance than the unsupervised/semi-supervised ones.
Cluster CCA (CCCA) \cite{ccca} develops associations between groups of data from the different modalities to appropriately design the common domain and can even handle unpaired data.
Generalized Multiview Analysis (GMA) \cite{gma} designs an effective algorithm to work with multi-domains even in the case of unseen classes.
Sparse representation based dictionary learning approaches have also been proposed for this problem in GCDL \cite{gcdl} and S2CDL \cite{s2cdl}.
To mitigate the difference between the common representations of the two modalities, adversarial loss along with classification loss and margin based triplet loss has been used in ACMR \cite{acmr} to perform cross-modal retrieval.
The work in \cite{sscmlp} can work under semi-supervised setting for both single and multi-label data.
It consists of a label prediction stage for the unlabeled data and a common domain representation learning stage to perform cross-modal retrieval.
Hashing based approaches like GSPH \cite{gsph}, SEPH \cite{seph}, 
GrowBit \cite{growbit} projects the data into the common hash bit representation domain and uses multiple techniques to preserve the semantic information provided by the labels.
The hashing approaches \cite{gsph} \cite{seph} \cite{growbit} also generate an unified representation for the data (provided both modality information are available) and generally shows better performance.
In addition, the work in \cite{gsph} \cite{growbit} shows strategies by which the algorithm can be scaled to any amount of training data.

\textbf{Incremental} methods for classification to handle data from new category or new task is a well studied problem and seminal approaches have been proposed in~\cite{icarl}\cite{lwf}\cite{lifelong}\cite{rebalance_cvpr2019}\cite{incremental_cvpr2019}\cite{lsil_cvpr2019}.
%Incremental learning problem can be divided into \textit{multi-class} and \textit{multi-task} setting \cite{rebalance_cvpr2019}.
The work in \cite{lwf} assumes that no data of the old categories are available and tries to overcome catastrophic forgetting using knowledge distillation.
\cite{icarl} formulates the objective as a nearest neighbor classification problem and utilises a small set of data from old categories (exemplars) to prevent forgetting through knowledge distillation.
%\cite{icarl} also discusses interesting strategies to construct the exemplar set and remove elements from it under maximum storage constraints.
The work in~\cite{lsil_cvpr2019} studies the incremental setting from the perspective of data imbalance (as the exemplar set is relatively small) and prediction bias towards the new categories and uses a balanced validation set and a bias correction layer to mitigate these issues.
Replacing the standard softmax classification with a cosine normalization layer for classification along with distillation and margin loss to prevent catastrophic forgetting is proposed in~\cite{rebalance_cvpr2019}.
An interesting approach to handle incremental setting in the hashing domain for image retrieval is proposed in \cite{incremental_cvpr2019}, where the hash codes for the old categories data are preserved and the hash codes for the new data are learned.
%The work in \cite{incremental_cvpr2019} has focused on one stage of incremental learning.
%Knowledge distillation in \cite{incremental_cvpr2019} is done by preserving the old hash codes and data (exemplar set) while learning the hash codes for the new categories.

{\bf Close to our work}
To the best of our knowledge the work in \cite{lifelong} is the only prior work for cross-modal retrieval in an incremental setting where different datasets are made sequentially available to train the common domain representation.
\cite{lifelong} preserves the old category information through knowledge distillation and without the use of any exemplars and performs the cross-modal evaluations on each dataset separately (\textit{multi-task} setting).

Our work has similarities with \cite{gsph} \cite{growbit} \cite{lifelong} but there are significant differences, which we will explain later.

Our work focuses on designing an incremental cross-modal algorithm for both retrieval and hashing applications in the \textit{multi-class} setting and takes inspirations from the work in \cite{lifelong} \cite{gsph} \cite{growbit} \cite{lsil_cvpr2019} \cite{incremental_cvpr2019} with subtle differences.
We have analyzed both a non-deep and deep based approach for our task.
As compared to the work in \cite{incremental_cvpr2019} which focuses on image retrieval our method works on cross-modal data.
In addition, we use distillation loss, weighted cross-entropy loss and imbalanced data sampling strategies to mitigate catastrophic forgetting across multiple phases of incoming data of new categories as compared to the work in \cite{incremental_cvpr2019} where only a single phase of new category data was considered.
Furthermore, the work in \cite{incremental_cvpr2019} uses an end-to-end strategy to learn the hash code and hash function together whereas we follow a two stage approach which was found to be much easier to solve and scales better for large amounts of training data \cite{gsph} \cite{growbit}.
As compared to the work in \cite{lifelong} which did retrieval under the \textit{multi-task} setting we focus on both retrieval and hashing under the more challenging \textit{multi-class} setting which is typically a much harder task \cite{rebalance_cvpr2019}.
In addition, as our common domain representation is selected to the hash bit representation, unification strategies \cite{seph} \cite{gsph} \cite{growbit} can be used to give much better performance in the hashing evaluation paradigm as evident from the later experiments.

% ***********************************************************************************   

\section{Problem Definition and Motivation}

Let us denote the original training data in the two modalities as $X^{tr} \in \mathcal{R}^{N^{tr} \times d_x}$ and $Y^{tr} \in \mathcal{R}^{N^{tr} \times d_y}$, where, $N^{tr}$ is the number of data samples and $d_x, d_y$ is the dimension of the data representation in their respective modalities (in general $d_x \neq d_y$).
Let the label information be denoted as $L^{tr} \in \mathcal{R}^{N^{tr} \times C}$ where $C$ is the number of categories.
Since we are dealing with single label data, each vector $\{ l^{tr}_i \}_{i=1}^{N^{tr}}$ is an one-hot representation of the label.
Given this initial training data, a cross-modal retrieval model (termed as {\em base model} from now) can be learnt using existing techniques. 
In the standard cross-modal retrieval paradigm, if the testing data is denoted by $X^{te}$ and $Y^{te}$, given a query from $X^{te}$ (or $Y^{te}$) modality, the objective is to retrieve semantically similar data from the search set i.e., $Y^{te}$ (or $X^{te}$) using the learnt model.

In this work, we address the problem of incremental cross-modal retrieval and hashing, where data from new categories are received in a sequential manner (or in batches) after the initial base model is learnt.
Let us denote the new set of training data as $\{ \hat{X}^{tr}, \hat{Y}^{tr}, \hat{L}^{tr} \}$, where the additional data consisting of $\hat{N}$ samples is spread over $\{ C+1, ..., C+\hat{C} \}$ categories i.e., over new $\hat{C}$ categories.
In the incremental paradigm, during testing, a query from either $X^{te}$ (or $Y^{te}$) might either come from the old $C$ or the new $\hat{C}$ category set.
Then, as data from new categories arise, the base model needs to be updated to learn about the new categories, without forgetting about the existing ones. \\ \\ 
% ***********************************************************************************
{\bf Motivation:} To the best of our knowledge, the multi-class incremental paradigm under cross-modal scenario has not been explored in literature.
So to understand the impact of subsequent stages of incremental learning on cross-modal retrieval, we perform a simple experiment with two non-deep (GMA \cite{gma}, SEPH \cite{seph}) and two deep approaches (ACMR \cite{acmr}, GrowBit \cite{growbit})  on two datasets (one containing image-text data - Wiki \cite{wiki} and the other having RGB images and 3D depth data - RGBD Object Database \cite{rgbd}).
\begin{figure}[t!]
	\begin{center}        
		\includegraphics[width=11.0cm,height=6.5cm]{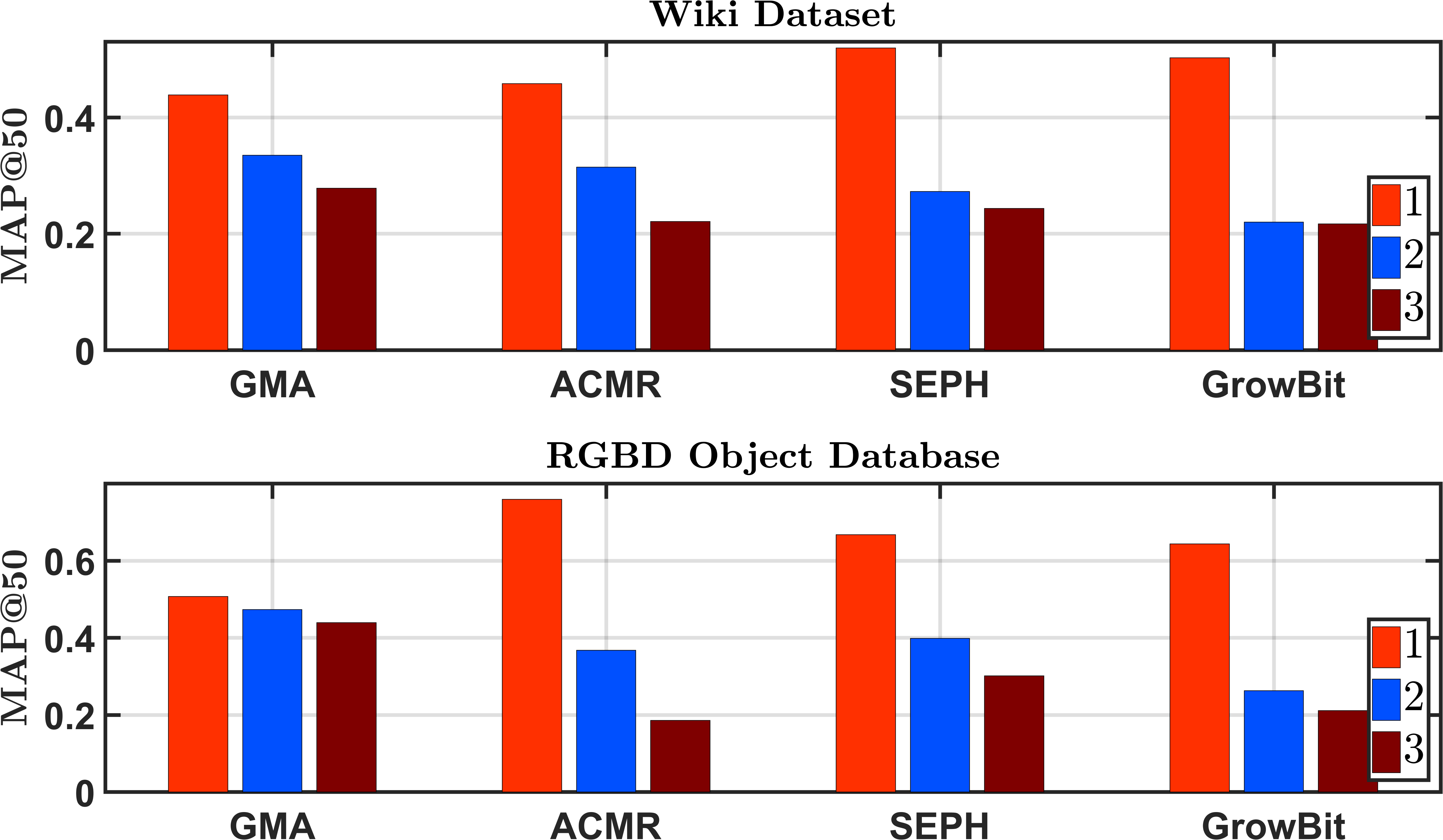}
	\end{center}
	\caption{Cross-Modal retrieval results (in MAP@50) for the Wiki \cite{wiki} (top) and RGBD Object Database \cite{rgbd} (bottom) when evaluated on the last training phase in the multi-class setting. The significant performance drop illustrates the need for using exemplars to retain knowledge of old category data and using techniques and tricks to update and adapt the trained model to reduce the performance drop.}
	\label{incremental_motivation}
\end{figure}
Figure \ref{incremental_motivation} shows the retrieval performance for 3 cases: (1) if the approach was trained with all the old and new data (best possible performance); (2) if the approach is trained with all the new data and few available data of the older classes and (3) the already learnt old model is kept fixed and used to perform the retrieval (worst possible performance).
The different results are marked as '1', '2' and '3' in Figure \ref{incremental_motivation}.
More details about the experiments are provided later in the Experiments section.
On each subsequent stage, more data from previously unseen categories are added for training and cross-modal retrieval evaluation is performed in the multi-class setting.
For the sake of clarity, we have reported the evaluation results after the last phase (when the data of all the categories have been revealed in an incremental fashion).
We observe that there is significant performance difference between the best and worst performance.
Using both the new and few old data samples does improve the performance, but this simple approach has two limitations, namely (1) there is still a lot of scope for improvement and (2) the model is still trained from scratch, which is inefficient since it is not utilizing the already trained model.
Next, we will describe the proposed approach for addressing this problem.

% ***********************************************************************************

\section{Proposed Approach}
The final objective is to compare data samples across two modalities with different representations, so the data needs to be projected to a common latent space where the relationships as given by their labels are preserved.
Inspired by~\cite{cvpr17}\cite{gsph}\cite{growbit}\cite{incremental_cvpr2019}, we use a two stage solution for this problem: (1) hash code learning and (2) hash function learning, which are performed sequentially.
%In the next stage, we learn a hash function to project the data $X^{tr}, Y^{tr}$ into the hash codes $A, B$.
In this work, we consider the binary representation $\{-1,1\}^q$ to be the common representation space, where $q$ is the hash code length.
Next we will look at the two steps in details.

% ***********************************************************************************
\subsection{Incremental Hash Code Learning}
In this stage, the hash codes $\{A, B\}$ for the data samples in $X^{tr}, Y^{tr}$ are learned which satisfies the semantic similarity as revealed by their labels $L^{tr}$.
First, we describe the base model~\cite{cvpr17}\cite{gsph}\cite{growbit}\cite{incremental_cvpr2019} briefly and then we describe the modification required for the incremental setting. \\ \\
% ***********************************************************************************
{\bf Base hash code learning: }
We first compute the similarity $S \in \{0,1\}^{N \times N}$ between the training data using $L^{tr}$.
Here, $S_{ij} = 1$ denotes data from same category and $S_{ij} = 0$ denotes data from different categories.
The following problem is solved to learn the hash codes 
\begin{eqnarray}
& & \min_{A, B} \vert \vert S - (1/q) A B^T \vert \vert_F^2 \nonumber \\
& & s.t. \hspace{5 pt} A \in \{-1,1\}^{N \times q}, B \in \{-1,1\}^{N \times q}
\label{eq_1}
\end{eqnarray}
Since it is difficult to solve \eqref{eq_1} over the discrete set $\{-1,1\}$, it is solved over the continuous relaxed set $[-1,1]$ to get an approximate solution.
In addition, if the training data is paired, an additional constraint can be used which forces the hash codes $A$ and $B$ to be as similar as possible.
Thus the final objective to solve for the hash codes is
\begin{eqnarray}
& & \min_{A, B} \vert \vert S - (1/q) A B^T \vert \vert_F^2 + \lambda_h \vert \vert A - B \vert \vert_F^2 \nonumber \\
& & s.t. \hspace{5 pt} A \in [-1,1]^{N \times q}, B \in [-1,1]^{N \times q}
\label{eq_2}
\end{eqnarray}
The regularization parameter $\lambda_h$ forces the hash codes of the two modalities to be close.
%The objective formulation in \eqref{eq_2} can be solved by (1) element-wise update (2) column-wise update or (3) matrix-wise update as shown in \cite{gsph} \cite{growbit}.
We use the computationally efficient matrix based method \cite{gsph} \cite{growbit} to solve for the hash codes.
Specifically, the gradients of the objective ($F$) in \eqref{eq_2} with respect to $A$ and $B$ can be computed as
\begin{eqnarray}
\nabla_A F & = & 2 A B^T B - 2 q S B + 2 \lambda_h (A - B) \nonumber \\
\nabla_A B & = & 2 B A^T A - 2 q S^T A - 2 \lambda_h (A - B)
\label{eq_3}
\end{eqnarray}
For this, the projected gradient descent algorithm is used and the matrices $A$ and $B$ are updated alternatively until convergence.
\begin{eqnarray}
A & = & \text{Proj}_{[-1,1]} (A - \eta_A \nabla_A F) \nonumber \\
B & = & \text{Proj}_{[-1,1]} (B - \eta_B \nabla_B F) 
\label{eq_4}
\end{eqnarray}
The Proj operation bounds the value of the matrix between $[-1,1]$.
A suitable value of the learning rate $\eta_A, \eta_B$ can be found out heuristically or by following the strategy in \cite{gsph}. \\ \\
% ***********************************************************************************
{\bf Learning hash codes for new categories:}
Here, we explain the proposed approach for updating the base model to reflect the new category data, while retaining information regarding how to deal with the old category data.
In the cross-modal retrieval scenario, the training data for all the old categories are not retained whereas, in the hashing scenario it is retained.
Following the same standard practice as used for building incremental image classification models ~\cite{icarl} \cite{lsil_cvpr2019} \cite{incremental_cvpr2019} \cite{rebalance_cvpr2019} we propose to use a few data samples of the old classes (termed as exemplars) to subsequently update and adapt our trained model.
Though different strategies can be used to select the best exemplars~\cite{icarl} \cite{lsil_cvpr2019} \cite{incremental_cvpr2019}, here, we randomly select $N^{samples}$ data (with paired correspondences) per old category $C$ to construct the exemplar set $\{ X^{e}, Y^{e}, L^{e} \}$ from $\{ X^{tr}, Y^{tr}, L^{tr} \}$.
Thus the cardinality of the exemplar set is $N^e = N^{samples} \times C$.
%The cardinality of the exemplar set $N^e = |X^{e}|=|Y^{e}|=|L^{e}|=N^{samples} \times C$.
We also retain the already learnt hash codes of the exemplar set $\{ A^{e}, B^{e} \}$ and the old hash functions $F_X^{old} = F_X$ and $F_Y^{old} = F_Y$.

Now the new training data is given by $\bar{X}^{tr} = [X^e \hspace{5 pt} \hat{X}^{tr}] \in \mathcal{R}^{\bar{N} \times d_x}$, $\bar{Y}^{tr} = [Y^e \hspace{5 pt} \hat{Y}^{tr}]  \in \mathcal{R}^{\bar{N}  \times d_y}$, $\bar{L}^{tr} = [L^e \hspace{5 pt} \hat{L}^{tr}]  \in \mathcal{R}^{\bar{N} \times (C + \hat{C})}$.
The size of this new training data is $\bar{N} = N^e+ \hat{N}$.
First, the new similarity matrix $\bar{S} \in \{0,1\}^{\bar{N} \times \bar{N}}$ is generated using the provided label information in $\bar{L}^{tr}$.
The next step is to learn the hash codes for the data in $\bar{X}^{tr}, \bar{Y}^{tr}$.
For this, we fix the hash codes for the old exemplar part $\{ A^{e}, B^{e} \}$ to retain the knowledge of the previous phase and only learn the hash codes of $\{ \hat{X}^{tr}, \hat{Y}^{tr} \}$ i.e., $\{ \hat{A}, \hat{B} \}$ from scratch.
This is analogous to the knowledge distillation loss used in incremental classification.
We can formulate the objective as 
\begin{eqnarray}
& & \min_{\bar{A}, \bar{B}} \vert \vert \bar{S} - (1/q) \bar{A} \bar{B}^T \vert \vert_F^2 + \lambda_h \vert \vert \bar{A} - \bar{B} \vert \vert_F^2 \nonumber \\
& & s.t. \hspace{5 pt} \bar{A} \in [-1,1]^{\bar{N} \times q}, \bar{B} \in [-1,1]^{\bar{N} \times q}
\label{eq_9}
\end{eqnarray}
which can be further decomposed (by considering a fixed $\{ A^{e}, B^{e} \}$) to 
\begin{eqnarray}
& & \min_{\hat{A}, \hat{B}} \vert \vert \bar{S} - (1/q) [A^e \hspace{5 pt} \hat{A}] [B^e \hspace{5 pt} \hat{B}]^T \vert \vert_F^2 + \lambda_h \vert \vert \hat{A} - \hat{B} \vert \vert_F^2 \nonumber \\
& & s.t. \hspace{5 pt} \hat{A} \in [-1,1]^{\hat{N} \times q}, \hat{B} \in [-1,1]^{\hat{N} \times q}
\label{eq_10}
\end{eqnarray}
Keeping the old hash codes fixed $\{ A^{e}, B^{e} \}$, the new hash codes $\{ \hat{A}, \hat{B} \}$ for the new category data in \eqref{eq_10} can be found by following the strategy in \cite{gsph}.
Thus the total hash codes for the exemplar and new category data is given by $\bar{A} = [A^e \hspace{5 pt} \hat{A}]$ and $\bar{B} = [B^e \hspace{5 pt} \hat{B}] $.
Thus, hash codes for the new data can be learnt without utilizing all the data samples of the previous training data.

This way of computing the new hash codes is similar to that used by~\cite{gsph} with the following differences.
(1) \cite{gsph} addresses the problem of online training when new data from the same categories are received, whereas here the incoming data can belong to completely new categories;
(2) \cite{gsph} uses the fixed batch to maintain the consistency of the learned hash codes between the small mini-batches and the entire training dataset, whereas here, we retain equal number of samples from each of the old categories to understand and maintain the relations of the new categories with the older ones.
Similar strategy to learn incremental hash codes in an end-to-end manner have been used in \cite{incremental_cvpr2019} for single modality image retrieval application.
The next step is to use the learnt hash codes to learn the hash functions.
%Next, we investigate how to adapt and update the hash functions $\{ F_X^{old}, F_Y^{old} \}$ to handle the new data.
% ***********************************************************************************

\subsection{Incremental Hash Function Learning}

Once the hash code has been learned, we need to learn the hash function $F_X : X \rightarrow A$ and $F_Y : Y \rightarrow B$ for computing the binary representation of the test samples.
Various strategies can be used to learn the hash function \cite{cvpr17} \cite{gsph} \cite{growbit} \cite{incremental_cvpr2019} like linear ridge regression, kernel logistic regression, deep neural networks, etc.
In this work, we focus on two strategies, namely linear ridge regression and deep neural network and show how they can be adapted to handle new incoming data.

We update the hash functions by taking into account the following two aspects: \\
1) The first objective is that the output of the new hash functions should be as close as possible to the original output for the existing data. 
This ensures that the already learnt information is not removed in the process of learning for the new categories (i.e., overcoming catastrophic forgetting). \\
2) Since the training data for updating the hash functions contains the exemplar set with few samples and the new training data with considerably more samples, it is highly imbalanced, which may affect the performance adversely.

% ***********************************************************************************
\subsubsection{Linear Ridge Regression}
First, we describe the base hash function learning and then the proposed modification. \\ \\
{\bf Base Hash Function Learning:} 
The base hash function learning is used after we obtain the hash codes for the base classes.
Here, both $F_X \in \mathcal{R}^{d_x \times q} = \{f_x^l\}_{l=1}^q$ and $F_Y \in \mathcal{R}^{d_y \times q} = \{f_y^l\}_{l=1}^q$ are learned for each of the $q$ bits separately.
The objective to solve can be written as 
\begin{eqnarray}
f_x^l & = & \min_{u_x} \vert \vert A_{*l} - X^{tr} u_x \vert \vert_2^2 + \lambda_{lr,x} \vert \vert u_x \vert \vert_2^2 \nonumber \\ 
f_y^l & = & \min_{u_y} \vert \vert B_{*l} - Y^{tr} u_y \vert \vert_2^2 + \lambda_{lr,y} \vert \vert u_y \vert \vert_2^2
\label{eq_5}
\end{eqnarray}
where, $u_x, u_y$ are the individual projection vectors to be learned, $A_{*l}, B_{*l}$ denote the $l^{th}$ column data (hash code) of the two matrix $A, B$ and $\lambda_{lr,x}, \lambda_{lr,y}$ are the regularization parameters  set by five-fold cross-validation experiments.
The objective in \eqref{eq_5} has a closed form solution. \\ \\
% ***********************************************************************************
% ***********************************************************************************
{\bf Updating the Linear Ridge Regression Functions:}
When data from new categories are obtained, these learnt hash functions must be updated to account for the new data also.
To update the hash functions from $F_X^{old} \rightarrow F_X$ and $F_Y^{old} \rightarrow F_Y$ we use of the training data $\{ \bar{X}^{tr}, \bar{Y}^{tr}, \bar{A}, \bar{B} \}$, which consists of the new data and the old exemplars.

To learn the hash functions $F_X, F_Y$ such that it learns for the new categories as well as retain the information from the previous phase operation, we propose three different models as given by the following objective functions (shown for the $X$ domain)
\begin{eqnarray}
f_x^l & = & \min_{u_x} \vert \vert \bar{A}_{*l} - \bar{X}^{tr} u_x \vert \vert_2^2 + \lambda_{lr,x} \vert \vert u_x \vert \vert_2^2 + \gamma_{x} \vert \vert u_x - f_x^{old,l} \vert \vert_2^2 \\ 
f_x^l & = & \min_{u_x} \vert \vert \bar{A}_{*l} - \bar{X}^{tr} u_x \vert \vert_2^2 + \lambda_{lr,x} \vert \vert u_x \vert \vert_2^2 + \gamma_{x} \vert \vert \bar{X}^{tr} u_x - \bar{X}^{tr} f_x^{old,l} \vert \vert_2^2 \\ 
f_x^l & = & \min_{u_x} \vert \vert \bar{A}_{*l} - \bar{X}^{tr} u_x \vert \vert_2^2 + \lambda_{lr,x} \vert \vert u_x \vert \vert_2^2 + \gamma_{x} \vert \vert u_x - f_x^{old,l} \vert \vert_2^2 + \gamma_{x} \vert \vert \bar{X}^{tr} u_x - \bar{X}^{tr} f_x^{old,l} \vert \vert_2^2
\end{eqnarray}
where, $f_x^{old,l}$ are the $l^{th}$ bit projection vector of the previous phase and $\gamma_x$ is a regularization parameter.
The first objective says that the new projection vector $f_x^l$ should be as close to the old projection vector $f_x^{old,l}$ as possible while simultaneously learning the regression function.
The second objective function says that the output of $f_x^l$ should be close to the output of the $f_x^{old,l}$ on the training data.
The third objective is a combination of the above two objectives.
Each of the objectives above has a closed form solution.
We can formulate similar objectives to update $F_Y^{old} \rightarrow F_Y$.

To account for the data imbalance between the old and new classes, we appropriately set the regularization parameters $\lambda_{lr,x}$ and $\gamma_x$.
In this work, they are set by five-fold cross-validation for which we construct a balanced validation set with equal amount of data per class from each of the exemplar classes $C$ and the new category set $\hat{C}$~\cite{lsil_cvpr2019}.
We have observed that this step plays a significant role in improving the overall performance by considerably mitigating the affect of data imbalance as also seen in the work in ~\cite{lsil_cvpr2019}.

% ***********************************************************************************
\subsubsection{Deep Neural Network}
Instead of using the linear ridge regression technique to learn the hash functions, we can similarly use a deep neural network architecture to learn $F_X$ and $F_Y$. \\ \\
{\bf Base Hash Function Learning:} 
Given the training data for the base classes, we use two multilayer neural networks to model $F_X$ and $F_Y$.
Each network has two fully connected (fc) layers with the configuration : $fc_1^t-rl-dr-fc_2^t-rl-dr$, where $rl$ and $dr$ denotes relu activation and dropout respectively.
The output of the $fc_2^t$ layer is sent to two separate $fc$ layers $fc_h^t$ and $fc_{ce}^t$.
($t$ denotes \{$X,Y$\}).
$fc_h^t$ outputs hash codes and hence requires the tanh activation, whereas $fc_{ce}^t$ predicts the class of the data sample and hence has Softmax activation.
Let us denote the data $X^{tr}$ having passed through $fc_1^X, fc_2^X$ as $X^{tr,lat}$ i.e., the latent representation, which is further passed through $fc_h^X$ to get the hash code $X^{tr,h}$ and through $fc_{ce}^X$ to get the discriminative class representation $X^{tr,ce}$.
The two losses used to train the base hash functions are defined below. \\ \\
% ***********************************************************************************
{\bf \textit{Hash Loss:} }The hash loss $\mathcal{L}^h$ makes the hash code output of the network $X^{tr,h}$ (or $Y^{tr,h}$) close to the expected hash code $A$ (or $B$) as is given by
%We utilize the following mean square loss function to enforce this
\begin{eqnarray}
\mathcal{L}^h = \sum_{i=1}^{N} \left(  \sum_{j=1}^{q} \left[  ( X^{tr,h}_{ij}-A_{ij} )^2 + ( Y^{tr,h}_{ij}-B_{ij} )^2 \right] \right)
\label{eq_6}
\end{eqnarray}
where, $i$ and $j$ denote $(i,j)^{th}$ element of the respective matrices. \\ \\
{\bf \textit{Classification Loss:}} The classification loss $\mathcal{L}^{ce}$ makes the latent representation output of the two networks $X^{tr,lat}, Y^{tr,lat}$ discriminative with respect to the provided labels.
%Hence the latent codes are further fed through the $fc_{ce}^t$ layer to generate the logits.
For this objective, we use the standard cross-entropy loss as 
\begin{eqnarray}
\mathcal{L}^{ce} = - \sum_{i=1}^{N} \left(  \sum_{j=1}^{C} \left[  L^{tr}_{ij} \log X^{tr,ce}_{ij} + L^{tr}_{ij} \log Y^{tr,ce}_{ij} \right] \right)
\label{eq_7}
\end{eqnarray}
The final loss $\mathcal{L}^{total}$ is thus given by $\mathcal{L}^{total} = \mathcal{L}^h + \mathcal{L}^{ce}$.
%\begin{eqnarray}
%\mathcal{L}^{total} = \mathcal{L}^h + \mathcal{L}^{ce}
%\label{eq_8}
%\end{eqnarray}
Once the hash functions $F_X$ and $F_Y$ are trained we can use it generate the hash codes for the testing data as $X^{te,h} = sign(F_X(X^{te})$) and $Y^{te,h} = sign(F_Y(Y^{te})$).
The $sign(.)$ operation is defined as $sign(x_{ij})=1$ if $x_{ij} \geq 0$ or $sign(x_{ij})=-1$ otherwise.
We solve the final objective $\mathcal{L}^{total}$ by using the standard stochastic gradient descent algorithm. \\ \\
% ***********************************************************************************
{\bf Updating the Deep Neural Network Functions:}
Given the original training data (or the current phase of training data), the deep neural network is trained as described above as is denoted as $F_X^{old}, F_Y^{old}$.
When data from new categories are obtained, two new neural network models denoted as $F_X, F_Y$ with similar architecture as $F_X^{old}, F_Y^{old}$ are instantiated.
The main difference is that the $fc_{ce}^t$ layer in the previous phase had a dimension of $C$, which is expanded to $C + \hat{C}$ for the new model.

To retain knowledge about the old category data, we copy the weights and biases of each layer (except that of the $fc_{ce}^t$ layer) in $F_X^{old}, F_Y^{old}$ to $F_X, F_Y$.
For the $fc_{ce}^t$ layer, we copy the weights only up to the $C$ category from the old to the new model.
To mitigate the effect of this imbalance on the learnt networks during training, we replace the loss in \eqref{eq_7} with the weighted cross entropy loss as
\begin{eqnarray}
\mathcal{L}^{wce} = - \sum_{i=1}^{N} \left(  \sum_{j=1}^{C} w_j \left[  L^{tr}_{ij} \log X^{tr,ce}_{ij} + L^{tr}_{ij} \log Y^{tr,ce}_{ij} \right] \right) \nonumber
\end{eqnarray}
where, $W = \{w_j\}_{j=1}^{C+\hat{C}}$ is the class weight for the $j^{th}$ class.
We find that a simple strategy of determining the class weights as $w_j = \frac{\bar{N}}{n_j}$, where $n_j$ is the number of samples of class $j$ in $\{\bar{X}_tr\}$ and $\bar{N}$ is the total number of samples works well for our problem.
Thus class weight $w_j$ relates to inverse of frequency of occurrence of each class samples.
We then define the total loss to be $\mathcal{L}^{total} = \mathcal{L}^{h} + \mathcal{L}^{wce}$ to train our model.
In addition, to handle the class imbalance, we have also utilised an imbalanced dataset sampler \footnote{https://github.com/ufoym/imbalanced-dataset-sampler} to generate the mini-batches of data.
This helps to oversample the data from the less commonly occurring classes while under-sampling from the more commonly ones during the process of mini-batch generation.
We have also investigated several other losses for obtaining the mentioned objectives (which we will briefly explain later), but the simple approach described above performed the best.
We have tried using $\mathcal{L}^{wce}$ during the training of neural networks in the first phase also but did not find any significant improvements.

%the using the distillation loss as shown in \cite{lwf} \cite{icarl} \cite{lsil_cvpr2019} \cite{incremental_cvpr2019} \cite{rebalance_cvpr2019} \cite{lifelong} on top of these strategies for retaining the information from the old phase but this strategies did not give significant improvements to the cross-modal retrieval performance.
%So in this work, we have used only a class weighted cross entropy loss coupled with an imbalanced data sampler to mitigate catastrophic forgetting.

% ***********************************************************************************

\section{Experiments}

Here, we report the results of extensive experiments and analysis performed to evaluate the effectiveness of the proposed framework and compare it against the baseline algorithms. 
First, we describe the datasets and the feature representations used, followed by description of three different evaluation protocols. \\ \\ 
% ***********************************************************************************
%\subsection{Datasets and Evaluation Protocol}
{\bf Datasets:} We have used four cross-modal datasets namely - Wiki dataset \cite{wiki}, LabelMe dataset \cite{labelme}, RGBD Object Database \cite{rgbd} and Pascal Sentences Dataset \cite{pascal_dataset} for evaluation.

The \textbf{Wiki dataset} \cite{wiki} consist of textual data and images collected from the online featured articles of the Wikipedia website and have been split into ten different categories.
The total no of data collected is $2866$, and the training : testing sets are obtained by doing a $70\%:30\%$ split of the data samples per class.
We use the same feature representation as in \cite{gssl} ($4096$-dim CNN \cite{jia2014caffe} for images \& 100-dim word vector \cite{mikolov2013distributed} for texts).
% ***********************************************************************************

The \textbf{LabelMe dataset} \cite{labelme} consist of images and their publicly available annotations as corresponding textual data spread over $8$ different categories.
The whole dataset contains about $2688$ image-text pairs and we use available GIST features for images and Word Frequency vector representation for texts as in \cite{prl_2017}.
As in~\cite{prl_2017}, we use  a $70\%:30\%$ split of the data samples per class to construct the train and test sets.
% ***********************************************************************************

The \textbf{Pascal Sentence dataset} \cite{pascal_dataset} consists of images with each image being described by five different sentence descriptions.
The data is spread over $20$ different categories with each class having around $50$ examples.
$35$ examples per class are used for training, while the rest are used for testing.
We use the $2048$-dim CNN feature representation from ResNet-101~\cite{resnet} for images.
We extract the textual features using BERT \cite{bert} \cite{text_feat_extractor} for each of the five sentences per example and then average it out to get the mean textual feature representation.
% ***********************************************************************************

The \textbf{RGBD Object Database} \cite{rgbd} is a large dataset having RGB and their corresponding Depth images spread over $51$ different categories.
We have extracted the $7000$-dim features using~\cite{rgbd_features} and then performed principal component analysis to reduce it to $50$-dim for both RGB and Depth data.
For our experiments, we have randomly selected $200$ sampler per class to construct our dataset with a $70\%:30\%$ split to construct our train and testing data. \\ \\
% ***********************************************************************************
% ***********************************************************************************
{\bf Evaluation Protocol:}  The proposed framework is developed for the challenging incremental \textit{multi-class} scenario, where the data from the new categories are available in phases, but the testing is performed over all the existing (old and new) categories.
For this, we divide the training data of the Wiki, LabelMe, Pascal Sentences and RGBD Object Database \cite{rgbd} in multiple phases.
The $10, 8, 20, 51$ categories of these datasets are divided into the following phases - $\{3,4,3\}, \{3,2,3\}, \{5,5,5,5\}$ and $\{11,10,10,10,10\}$ respectively.
We shuffle the order of classes randomly three times, and {\bf results averaged over the three shuffling orders are reported.}
For evaluation, we have used three different protocols. \\ \\
{\bf Protocol P-I:} 
Here, the baseline model is trained using all the currently available date (the newest phase and all the old phases) to perform the testing, which is akin to throwing away the old trained model and retraining it from scratch.
Thus performance under P-I represents the upper-bound retrieval accuracy of each algorithm. \\ \\
% ***********************************************************************************
{\bf Protocol P-II:} 
Here, the baseline model is trained only using data from the first phase and is not updated using data from new categories.
Since testing is done without any adaptation, performance under  P-II represents the lower-bound retrieval accuracy of each algorithm. \\ \\
% ***********************************************************************************
{\bf Protocol P-III:} 
Here, the knowledge of the data in the previous phases is retained through proper selection of exemplars.
Each baseline model is trained from scratch using the data of the new phase and the old exemplar data.
The proposed incremental model is trained in this setting by adapting from the baseline model using the proposed approach.

For evaluation, we report the Mean Average Precision (MAP) as used in \cite{gssl} \cite{gsph}.
%MAP can be computed as the mean of the average precision of all queries AP which can be computed as $AP(q) = \frac{\sum_{r=1}^{R} P_q(r) \delta(r)}{\sum_{r=1}^{R} \delta(r)}$, where $R$ is the number of retrieved items and $P_q(r)$ is the precision at position $r$ for query $q$.
%$\delta(r)$ is set to be $1$ if the labels of the query and retrieved samples are the same else it is set to $0$.
We measure the MAP with respect to both image and textual queries and report the average MAP here.
We report MAP@50 for cross-modal retrieval experiments and MAP@all for hashing experiments.
For cross-modal retrieval, the gallery data is revealed during the testing time and hence their labels are not known.
For cross-modal hashing, the training data is the same as gallery data and hence unification strategies or non-regenerated hash codes (wherever applicable as in \cite{seph} \cite{gsph} \cite{growbit}) can be used to obtain better performance.

For comparison, we have considered the following standard cross-modal algorithms: 
(1) Non-deep approaches - CCCA \cite{ccca}, GMA \cite{gma}, GSSL \cite{gssl}, GSPH \cite{gsph}, SEPH \cite{seph} and 
(2) Deep approaches - GrowBit \cite{growbit}, ACMR \cite{acmr} and SSCMLP \cite{sscmlp}.
We have used the publicly available implementation of all the algorithms and reimplemented ACMR ourselves for our evaluation.
%We also included the GrowBit algorithm with an additional classification loss for GrowBit$_{ce}$ for our experiments.
We have evaluated all our hashing algorithms on hash code size of $128$ bits.

% **********************************************************************************
\subsection{Evaluation of baselines under P-I and P-II}
Here, we evaluate the standard baseline algorithms under protocols \textbf{P-I} and \textbf{P-II} and report the Average MAP@50 and MAP@all results for the retrieval and hashing experiments.
Results for Wiki and LabelMe are reported in Table~\ref{me}, and that for Pascal Sentence and RGBD Object databases are reported in Tables~\ref{pascal_results} and~\ref{rgbd_results} respectively.
For the hashing results (denoted with ``$*$"), for \textbf{P-I}, we use the hash codes of the training data itself during testing (since the gallery and training sets are same).
Under \textbf{P-II}, during Phase-1, we use the gallery hash codes generated during training, and for the other phases, the unified hash codes for the gallery data are generated to get the best performance.

For \textbf{P-I}, we train the baseline models from scratch using all the data available (for old and new categories) and thus represents the performance upper bound.
We make the following observations.
The deep methods usually perform better than their non-deep counterparts.
Also, the algorithms like GSPH \cite{gsph}, SEPH \cite{seph}, GrowBit \cite{growbit} perform better in the hashing paradigm as compared to the other algorithms due to their ability to use the original hash codes for the gallery data.
We also observe that as the incoming data is added in phases, the average MAP monotonically decreases across all the baselines and datasets.
This is due to the \textit{multi-class} testing scenario, where during testing, the query can come from the new as well as the old categories, which makes the problem more challenging.
% ***********************************************************************************************************

For \textbf{P-II}, the algorithms are trained only for Phase 1, and are not adapted for subsequent new data, and thus represents the performance lower bound.
We also observe from the results that the performance degradation as data from new phases are added are more severe as compared to \textbf{P-I} for both cross-modal retrieval and hashing.
Interestingly, for cross-modal retrieval, performance degradation of GMA \cite{gma} is much lesser with the addition of new data as compared to other algorithms.
We also observe that the performance of the non-deep baselines are lower than the deep based baseline algorithms for RGBD Object database.
% ***********************************************************************************************************

\begin{table*}[t!]
	\tiny
	\centering
	\renewcommand{\arraystretch}{1.4}
	\setlength{\tabcolsep}{9.8 pt}
	\caption{Average MAP@50 and MAP@all for Wiki and LabelMe datasets for the baseline algorithms under \textbf{P-I}, \textbf{P-II} and \textbf{P-III} protocols for cross-modal retrieval and hashing experiments (denoted by ``*"). We also report the performance of our algorithms ICMH - LR (all the 3 variants) and ICMH - Deep.
		%, which by updating the model parameters on each phase by using the exemplar data for the old categories and all the available data of the new categories.
		%``*" represents that the algorithm is working under the hashing scenario.
		The results are provided as \textbf{P-III} (\textbf{P-I}, \textbf{P-II}) with ``bar", ``underline" signifying the upper-bound and lower-bound performance for each phase.
		For phase 1, the results are obtained with the base model and is thus same for all protocols.}
	%The superior performance of our proposed models under both the cross-modal retrieval and hashing scenario is evident from the results.}
	\label{table_PIII}
	\begin{tabular}{|c|c|c|c|c|c|c|}
		\hline
		Algorithms & \multicolumn{3}{c|}{Wiki} & \multicolumn{3}{c|}{LabelMe} \\ \hline
		Phases & 1 & 2 & 3 & 1 & 2 & 3 \\ \hline \hline
		CCCA & 0.669 & 0.411 ($\overline{0.496}$, $\underline{0.305}$) & 0.290 ($\overline{0.407}$, $\underline{0.220}$) & 0.820 & 0.646 ($\overline{0.732}$, $\underline{0.434}$) & 0.436 ($\overline{0.602}$, $\underline{0.273}$) \\
		GMA & 0.666 & 0.434 ($\overline{0.502}$, $\underline{0.353}$) & 0.335 ($\overline{0.438}$, $\underline{0.278}$) & 0.859 & 0.696 ($\overline{0.797}$, $\underline{0.567}$) & 0.535 ($\overline{0.721}$, $\underline{0.376}$) \\
		GSSL & 0.681 & 0.414 ($\overline{0.530}$, $\underline{0.310}$) & 0.303 ($\overline{0.474}$, $\underline{0.228}$) & 0.903 & 0.676 ($\overline{0.862}$, $\underline{0.520}$) & 0.516 ($\overline{0.801}$, $\underline{0.318}$) \\
		SEPH & 0.687 & 0.386 ($\overline{0.578}$, $\underline{0.341}$) & 0.272 ($\overline{0.519}$, $\underline{0.243}$) & 0.909 & 0.522 ($\overline{0.861}$, $\underline{0.524}$) & 0.474 ($\overline{0.834}$, $\underline{0.339}$) \\
		GSPH$_{klr}$ & 0.696 & 0.371 ($\overline{0.574}$, $\underline{0.340}$) & 0.263 ($\overline{0.513}$, $\underline{0.255}$) & 0.911 & 0.543 ($\overline{0.872}$, $\underline{0.540}$) & 0.458 ($\overline{0.839}$, $\underline{0.357}$) \\
		GSPH$_{lin}$ & 0.664 & 0.401 ($\overline{0.533}$, $\underline{0.328}$) & 0.264 ($\overline{0.477}$, $\underline{0.254}$) & 0.892 & 0.540 ($\overline{0.851}$, $\underline{0.551}$) & 0.404 ($\overline{0.810}$, $\underline{0.361}$) \\
		ACMR & 0.718 & 0.432 ($\overline{0.530}$, $\underline{0.301}$) & 0.314 ($\overline{0.458}$, $\underline{0.221}$) & 0.932 & 0.724 ($\overline{0.902}$, $\underline{0.273}$) & 0.595 ($\overline{0.858}$, $\underline{0.197}$) \\
		GrowBit & 0.659 & 0.363 ($\overline{0.558}$, $\underline{0.306}$) & 0.220 ($\overline{0.502}$, $\underline{0.216}$) & 0.903 & 0.528 ($\overline{0.874}$, $\underline{0.498}$) & 0.349 ($\overline{0.839}$, $\underline{0.300}$) \\
		%GrowBit$_{ce}$ & 0.656 & 0.401 ($\overline{0.560}$, $\underline{0.296}$) & 0.263 ($\overline{0.493}$, $\underline{0.233}$) & 0.895 & 0.675 ($\overline{0.863}$, $\underline{0.527}$) & 0.546 ($\overline{0.837}$, $\underline{0.314}$) \\
		SSCMLP & 0.734 & 0.415 ($\overline{0.550}$, $\underline{0.291}$) & 0.287 ($\overline{0.482}$, $\underline{0.216}$) & 0.928 & 0.706 ($\overline{0.900}$, $\underline{0.302}$) & 0.565 ($\overline{0.496}$, $\underline{0.205}$) \\
		{\bf ICMH - LR$_{1;2;3}$} & 0.678 & 0.469/ 0.456/ 0.434 & {\bf 0.386}/ 0.373/ 0.359 & 0.906 & 0.700/ 0.740/ {\bf 0.770} & 0.613/ 0.628/ 0.671 \\
		{\bf ICMH - Deep} & 0.659 & {\bf 0.480} & 0.384 & 0.906 & 0.767 & {\bf 0.694} \\
		{\bf ICMH - Deep + $\mathcal{L}_d$} & 0.655 & 0.479 & 0.390 & 0.906 & 0.781 & 0.710 \\ \hline \hline
		% **********************************************************************************************
		CCCA$^*$ & 0.666 & 0.391 ($\overline{0.508}$, $\underline{0.259}$) & 0.238 ($\overline{0.378}$, $\underline{0.174}$) & 0.595 & 0.430 ($\overline{0.475}$, $\underline{0.336}$) & 0.287 ($\overline{0.357}$, $\underline{0.206}$) \\
		GMA$^*$ & 0.564 & 0.283 ($\overline{0.352}$, $\underline{0.239}$) & 0.193 ($\overline{0.280}$, $\underline{0.166}$) & 0.605 & 0.439 ($\overline{0.455}$, $\underline{0.359}$) & 0.278 ($\overline{0.343}$, $\underline{0.229}$) \\
		GSSL$^*$ & 0.753 & 0.386 ($\overline{0.539}$, $\underline{0.314}$)  & 0.229 ($\overline{0.449}$, $\underline{0.210}$) & 0.859 & 0.530 ($\overline{0.775}$, $\underline{0.498}$) & 0.377 ($\overline{0.663}$, $\underline{0.286}$) \\
		SEPH$^*$ & 0.829 & 0.528 ($\overline{0.578}$, $\underline{0.443}$)  & 0.372 ($\overline{0.519}$, $\underline{0.335}$) & 0.950 & 0.634 ($\overline{0.924}$, $\underline{0.638}$) & 0.536 ($\overline{0.895}$, $\underline{0.439}$) \\
		GSPH$_{klr}^*$ & 0.828 & 0.538 ($\overline{0.740}$, $\underline{0.401}$) & 0.373 ($\overline{0.694}$, $\underline{0.286}$) & 0.949 & 0.592 ($\overline{0.929}$, $\underline{0.599}$) & 0.504 ($\overline{0.899}$, $\underline{0.397}$) \\
		GSPH$_{lin}^*$ & 0.828 & 0.402 ($\overline{0.733}$, $\underline{0.212}$) & 0.313 ($\overline{0.680}$, $\underline{0.168}$) & 0.925 & 0.338 ($\overline{0.905}$, $\underline{0.251}$) & 0.430 ($\overline{0.877}$, $\underline{0.185}$) \\
		ACMR$^*$ & 0.823 & 0.493 ($\overline{0.679}$, $\underline{0.247}$) & 0.297 ($\overline{0.621}$, $\underline{0.164}$) & 0.949 & 0.648 ($\overline{0.923}$, $\underline{0.265}$) & 0.521 ($\overline{0.887}$, $\underline{0.173}$) \\
		GrowBit$^*$ & 0.843 & 0.423 ($\overline{0.737}$, $\underline{0.333}$) & 0.212 ($\overline{0.690}$, $\underline{0.207}$) & 0.952 & 0.454 ($\overline{0.927}$, $\underline{0.500}$) & 0.348 ($\overline{0.900}$, $\underline{0.271}$) \\
		%GrowBit$_{ce}^*$ & 0.840 & 0.452 ($\overline{0.739}$, $\underline{0.321}$) & 0.239 ($\overline{0.691}$, $\underline{0.206}$) & 0.953 & 0.583 ($\overline{0.928}$, $\underline{0.551}$) & 0.490 ($\overline{0.900}$, $\underline{0.288}$) \\
		SSCMLP$^*$ & 0.840 & 0.475 ($\overline{0.679}$, $\underline{0.258}$) & 0.272 ($\overline{0.600}$, $\underline{0.179}$) & 0.954 & 0.626 ($\overline{0.956}$, $\underline{0.284}$) & 0.484 ($\overline{0.927}$, $\underline{0.180}$) \\
		{\bf ICMH - LR$_{1;2;3}^*$} & 0.827 & {\bf 0.681}/0.671/0.674 & 0.620/{\bf 0.630}/0.628 & 0.944 & 0.795/0.824/0.849 & 0.756/0.762/{\bf 0.803} \\
		{\bf ICMH - Deep$^*$} & 0.837 & 0.674 & 0.595 & 0.951 & {\bf 0.853} & 0.795 \\
		{\bf ICMH - Deep + $\mathcal{L}_d^*$} & 0.837 & 0.675 & 0.588 & 0.952 & 0.858 & 0.801 \\ \hline
	\end{tabular}
	\label{me}
\end{table*}

% ***********************************************************************************************************

\begin{table*}[t!]
	\tiny
	\centering
	\renewcommand{\arraystretch}{1.4}
	\setlength{\tabcolsep}{18.8 pt}
	\caption{Average MAP@50 and MAP@all for Pascal Sentences dataset for the baseline algorithms under \textbf{P-I}, \textbf{P-II} and \textbf{P-III} protocols for cross-modal retrieval and hashing experiments (denoted by ``*"). We also report the performance of our algorithms ICMH - LR (all the 3 variants) and ICMH - Deep.
		%, which by updating the model parameters on each phase by using the exemplar data for the old categories and all the available data of the new categories.
		%``*" represents that the algorithm is working under the hashing scenario.
		The results are provided as \textbf{P-III} (\textbf{P-I}, \textbf{P-II}) with ``bar", ``underline" signifying the upper-bound and lower-bound performance for each phase.
		For phase 1, the results are obtained with the base model and is thus same for all protocols.}
	%The superior performance of our proposed models under both the cross-modal retrieval and hashing scenario is evident from the results.}
	\label{table_pascal}
	\begin{tabular}{|c|c|c|c|c|}
		\hline
		Algorithms & \multicolumn{4}{c|}{Pascal Sentences} \\ \hline
		Phases & 1 & 2 & 3 & 4 \\ \hline \hline
		CCCA & 0.634 & 0.460 ($\overline{0.537}$, $\underline{0.330}$) & 0.296 ($\overline{0.385}$, $\underline{0.231}$) & 0.267 ($\overline{0.306}$, $\underline{0.178}$) \\
		GMA & 0.650 & 0.547 ($\overline{0.594}$, $\underline{0.365}$) & 0.448 ($\overline{0.500}$, $\underline{0.267}$) & 0.371 ($\overline{0.421}$, $\underline{0.215}$) \\
		GSSL & 0.668 & 0.545 ($\overline{0.616}$, $\underline{0.343}$) & 0.432 ($\overline{0.536}$, $\underline{0.241}$) & 0.384 ($\overline{0.462}$, $\underline{0.192}$) \\
		SEPH & 0.669 & 0.529 ($\overline{0.641}$, $\underline{0.357}$)&  0.406 ($\overline{0.536}$, $\underline{0.250}$) & 0.349 ($\overline{0.446}$, $\underline{0.205}$) \\
		GSPH$_{klr}$ & 0.669 & 0.496 ($\overline{0.618}$, $\underline{0.333}$) & 0.389 ($\overline{0.549}$, $\underline{0.236}$) & 0.333 ($\overline{0.465}$, $\underline{0.193}$) \\
		GSPH$_{lin}$ & 0.657 & 0.536 ($\overline{0.640}$, $\underline{0.327}$) & 0.411 ($\overline{0.536}$, $\underline{0.226}$) & 0.364 ($\overline{0.478}$, $\underline{0.181}$) \\
		ACMR & 0.685 & 0.524 ($\overline{0.603}$, $\underline{0.257}$) & 0.409 ($\overline{0.505}$, $\underline{0.180}$) & 0.332 ($\overline{0.416}$, $\underline{0.152}$) \\
		GrowBit & 0.681 & 0.460 ($\overline{0.629}$, $\underline{0.346}$) & 0.347 ($\overline{0.548}$, $\underline{0.242}$) & 0.293 ($\overline{0.479}$, $\underline{0.195}$) \\
		%GrowBit$_{ce}$ & 0.684 & 0.504 ($\overline{0.627}$, $\underline{0.338}$) & 0.373 ($\overline{0.560}$, $\underline{0.248}$) & 0.322 ($\overline{0.475}$, $\underline{0.195}$) \\
		SSCMLP & 0.680 & 0.545 ($\overline{0.613}$, $\underline{0.283}$) & 0.413 ($\overline{0.531}$, $\underline{0.192}$) & 0.363 ($\overline{0.456}$, $\underline{0.151}$) \\
		{\bf ICMH - LR$_{1;2;3}$} & 0.665 & 0.574/0.515/0.511 & 0.501/0.443/0.462 & 0.424/0.355/0.398 \\
		{\bf ICMH - Deep} & 0.670 & {\bf 0.589} & {\bf 0.504} & {\bf 0.427} \\
		{\bf ICMH - Deep + $\mathcal{L}_d$} & 0.670 & 0.587 & 0.496 & 0.420 \\ \hline \hline
		% ***********************************************************************************************************
		CCCA$^*$ & 0.513 & 0.344 ($\overline{0.366}$, $\underline{0.277}$) & 0.211 ($\overline{0.248}$, $\underline{0.182}$) & 0.166 ($\overline{0.187}$, $\underline{0.136}$) \\
		GMA$^*$ & 0.692 & 0.466 ($\overline{0.611}$, $\underline{0.311}$) & 0.332 ($\overline{0.525}$, $\underline{0.202}$) & 0.243 ($\overline{0.451}$, $\underline{0.149}$) \\
		GSSL$^*$ & 0.715 & 0.544 ($\overline{0.643}$, $\underline{0.369}$) & 0.418 ($\overline{0.566}$, $\underline{0.247}$) & 0.329 ($\overline{0.503}$, $\underline{0.183}$) \\
		SEPH$^*$ & 0.814 & 0.661 ($\overline{0.788}$, $\underline{0.444}$) & 0.548 ($\overline{0.720}$, $\underline{0.309}$) & 0.485 ($\overline{0.654}$, $\underline{0.236}$) \\
		GSPH$_{klr}^*$ & 0.805 & 0.634 ($\overline{0.776}$, $\underline{0.414}$) & 0.543 ($\overline{0.288}$, $\underline{0.305}$) & 0.479 ($\overline{0.665}$, $\underline{0.219}$) \\
		GSPH$_{lin}^*$ & 0.813 & 0.516 ($\overline{0.794}$, $\underline{0.210}$) & 0.565 ($\overline{0.734}$, $\underline{0.154}$) & 0.495 ($\overline{0.686}$, $\underline{0.122}$) \\
		ACMR$^*$ & 0.811 & 0.616 ($\overline{0.765}$, $\underline{0.321}$) & 0.485 ($\overline{0.694}$, $\underline{0.219}$) & 0.392 ($\overline{0.615}$, $\underline{0.163}$) \\
		GrowBit$^*$ & 0.816 & 0.560 ($\overline{0.784}$, $\underline{0.411}$) & 0.432 ($\overline{0.725}$, $\underline{0.267}$) & 0.332 ($\overline{0.676}$, $\underline{0.200}$) \\
		%GrowBit$_{ce}^*$ & 0.817 & 0.607 ($\overline{0.777}$, $\underline{0.409}$) & 0.473 ($\overline{0.730}$, $\underline{0.268}$) & 0.399 ($\overline{0.666}$, $\underline{0.200}$) \\
		SSCMLP$^*$ & 0.812 & 0.569 ($\overline{0.765}$, $\underline{0.323}$) & 0.426 ($\overline{0.704}$, $\underline{0.216}$) & 0.345 ($\overline{0.640}$, $\underline{0.163}$) \\
		{\bf ICMH - LR$_{1;2;3}^*$} & 0.806 & 0.744/0.705/0.694 & {\bf 0.703}/0.664/0.676 & 0.637/0.589/0.619 \\
		{\bf ICMH - Deep$^*$} & 0.807 & {\bf 0.745} & 0.693 & {\bf 0.640} \\ 
		{\bf ICMH - Deep + $\mathcal{L}_d^*$} & 0.807 & 0.749 & 0.688 & 0.629 \\ \hline
	\end{tabular}
	\label{pascal_results}
\end{table*}

\begin{figure}[t!]
	\begin{center}        
		\includegraphics[width=12.2cm,height=8.5cm]{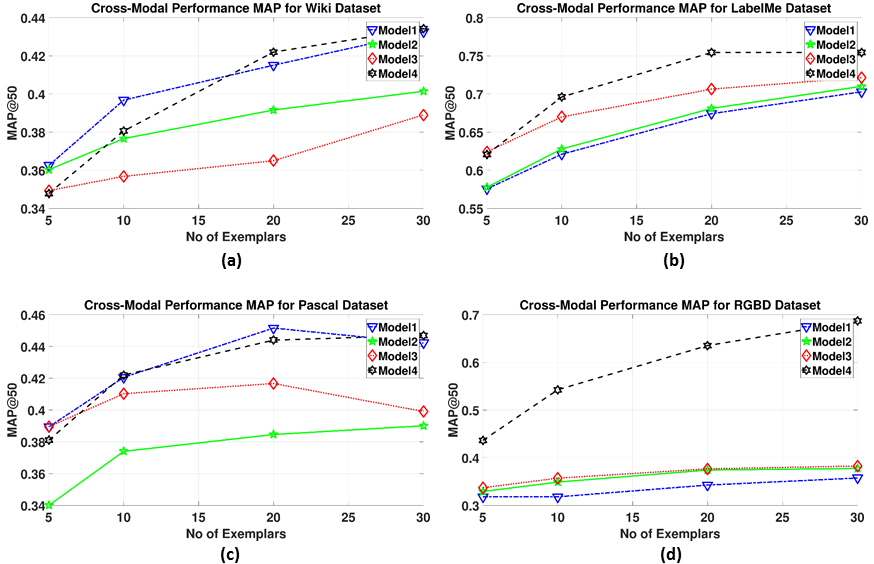}
	\end{center}
	\caption{Cross-modal retrieval performance in MAP@50 (average result of the final stage) in multi-class setting for the Wiki \cite{wiki}, LabelMe \cite{labelme}, Pascal Sentences \cite{pascal_dataset} and RGBD Object Database \cite{rgbd} with increase in the number of exemplars for adapting and updating the incremental model.}
	\label{cm_exemplars}
\end{figure}

\begin{figure}[t!]
	\begin{center}        
		\includegraphics[width=12.4cm,height=8.5cm]{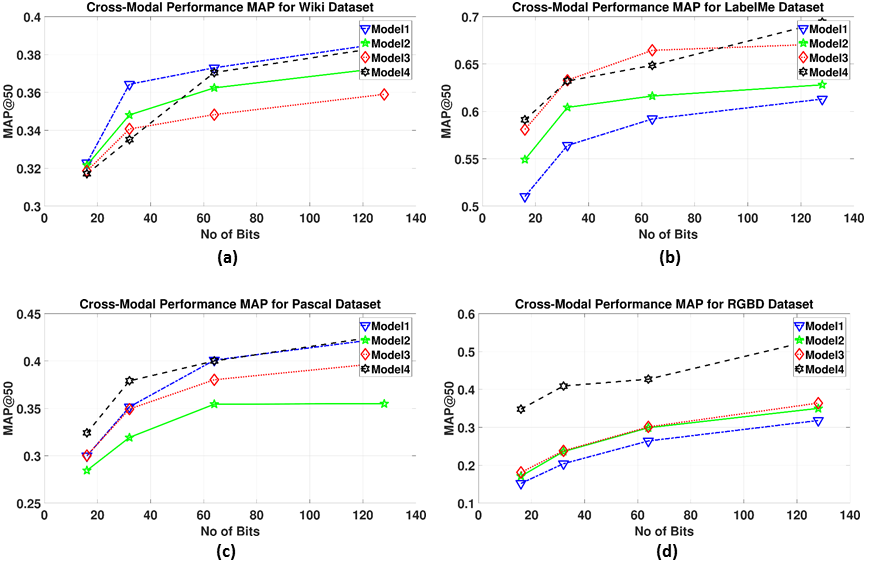}
	\end{center}
	\caption{Cross-modal retrieval performance in MAP@50 (average result of the final stage) in multi-class setting for the Wiki \cite{wiki}, LabelMe \cite{labelme}, Pascal Sentences \cite{pascal_dataset} and RGBD Object Database \cite{rgbd} with increase in the number of hash bits.}
	\label{cm_bits}
\end{figure}

% **********************************************************************************
\subsection{Evaluation and Comparisons under P-III}

From the previous section, we observe that when new data is added in phases, the model should be updated accordingly to mitigate significant performance degradation.
We first analyze the performance of the baseline algorithms when we retrain the model from scratch using the data from the current phase (over the new categories) and the small amount of data from the old categories (exemplars).
Finally we report the results of the proposed ICMH - LR (three different variants of the non-deep hash function adaptation based on Section 3.5.1) and ICMH - Deep in this setting. 
For these experiments, we have used the exemplar set per category to be $10$.

We observe from the results that the performance of the baseline models have mostly improved as compared to the results under \textbf{P-II}, since the model has seen the new category in addition to the a portion of the old category data.
However, the performance is significantly worse than under \textbf{P-I}, where the model had access to all the data of the old and new categories.
In addition, it is important to remember that the baseline models have been trained from scratch in \textbf{P-III} which makes it an inefficient approach.
The objective is to use the proposed techniques to bridge the performance gap between \textbf{P-I} and \textbf{P-II}.

We observe from the results that the performance of the proposed models is in between \textbf{P-I} and \textbf{P-II}.
%This is expected since we adapt our model to retain the old information (through exemplars) and also using weighted cross-entropy loss and an imbalanced data sampler.
For Wiki, LabelMe and Pascal Sentences data, both our non-deep and deep models are performing significantly better than the other baselines under \textbf{P-III}.
In general, ICMH - LR$_1$ performs consistently well as compared to the other two variations across all the datasets.
The performance of GMA \cite{gma} is found to be remarkably good under \textbf{P-III}, though no special adaptation scheme was used to update the model.
For the RGBD Object data, since the non-deep GSPH$_{lin}$ performs poorly as compared to the other baselines (as evident from the results in \textbf{P-I}), even though proposed adaptation strategies improved the results, it was not satisfactory.
However we observe significant improvement in the cross-modal retrieval and hashing accuracy when the proposed ICMH - Deep framework is used.
We have also reported the results of ICMH - Deep when added with the distillation loss $\mathcal{L}^d$ \cite{icarl} \cite{lwf} \cite{lifelong} but did not find any significant improvements.

\begin{table*}[t!]
	\tiny
	\centering
	\renewcommand{\arraystretch}{1.4}
	\setlength{\tabcolsep}{12.0 pt}
	\caption{Average MAP@50 and MAP@all for RGBD Object Database for the baseline algorithms under \textbf{P-I}, \textbf{P-II} and \textbf{P-III} protocols for cross-modal retrieval and hashing experiments (denoted by ``*"). We also report the performance of our algorithms ICMH - LR (all the 3 variants) and ICMH - Deep.
		%, which by updating the model parameters on each phase by using the exemplar data for the old categories and all the available data of the new categories.
		%``*" represents that the algorithm is working under the hashing scenario.
		The results are provided as \textbf{P-III} (\textbf{P-I}, \textbf{P-II}) with ``bar", ``underline" signifying the upper-bound and lower-bound performance for each phase.
		For phase 1, the results are obtained with the base model and is thus same for all protocols.}
	%The superior performance of our proposed models under both the cross-modal retrieval and hashing scenario is evident from the results.}
	\label{table_rgbd}
	\begin{tabular}{|c|c|c|c|c|c|}
		\hline
		Algorithms & \multicolumn{5}{c|}{RGBD Object Database} \\ \hline
		Phases & 1 & 2 & 3 & 4 & 5 \\ \hline \hline
		CCCA & 0.513 & 0.324 ($\overline{0.369}$, $\underline{0.327}$) & 0.231 ($\overline{0.277}$, $\underline{0.244}$) & 0.181 ($\overline{0.240}$, $\underline{0.196}$) & 0.159 ($\overline{0.207}$, $\underline{0.166}$) \\
		GMA & 0.697 & 0.601 ($\overline{0.635}$, $\underline{0.560}$) & 0.514 ($\overline{0.564}$, $\underline{0.490}$) & 0.493 ($\overline{0.529}$, $\underline{0.458}$) & 0.473 ($\overline{0.507}$, $\underline{0.439}$) \\
		GSSL & 0.764 & 0.473 ($\overline{0.667}$, $\underline{0.450}$) & 0.323 ($\overline{0.555}$, $\underline{0.327}$) & 0.274 ($\overline{0.487}$, $\underline{0.264}$) & 0.249 ($\overline{0.439}$, $\underline{0.226}$) \\
		SEPH & 0.946 & 0.665 ($\overline{0.895}$, $\underline{0.598}$) & 0.521 ($\overline{0.794}$, $\underline{0.442}$) & 0.460 ($\overline{0.724}$, $\underline{0.356}$) & 0.398 ($\overline{0.667}$, $\underline{0.301}$) \\
		GSPH$_{klr}$ & 0.946 & 0.613 ($\overline{0.897}$, $\underline{0.575}$) & 0.461 ($\overline{0.802}$, $\underline{0.431}$) & 0.396 ($\overline{0.736}$, $\underline{0.348}$) & 0.346 ($\overline{0.672}$, $\underline{0.295}$) \\
		GSPH$_{lin}$ & 0.831 & 0.518 ($\overline{0.696}$, $\underline{0.499}$) & 0.352 ($\overline{0.558}$, $\underline{0.371}$) & 0.292 ($\overline{0.470}$, $\underline{0.295}$) & 0.253 ($\overline{0.408}$, $\underline{0.253}$) \\
		ACMR & 0.916 & 0.377 ($\overline{0.916}$, $\underline{0.345}$) &  0.461 ($\overline{0.856}$, $\underline{0.271}$) & 0.408 ($\overline{0.812}$, $\underline{0.214}$) & 0.367 ($\overline{0.759}$, $\underline{0.186}$) \\
		GrowBit & 0.938 & 0.517 ($\overline{0.913}$, $\underline{0.475}$) & 0.348 ($\overline{0.828}$, $\underline{0.329}$) & 0.301 ($\overline{0.736}$, $\underline{0.256}$) & 0.263 ($\overline{0.643}$, $\underline{0.211}$) \\
		%GrowBit$_{ce}$ & 0.945 & 0.608 ($\overline{0.930}$, $\underline{0.486}$) & 0.478 ($\overline{0.877}$, $\underline{0.337}$) & 0.417 ($\overline{0.833}$, $\underline{0.262}$) & 0.373 ($\overline{0.787}$, $\underline{0.215}$) \\
		SSCMLP & 0.958 & 0.676 ($\overline{0.940}$, $\underline{0.393}$) & 0.625 ($\overline{0.892}$, $\underline{0.285}$) & 0.563 ($\overline{0.864}$, $\underline{0.223}$) & 0.518 ($\overline{0.842}$, $\underline{0.187}$) \\
		{\bf ICMH - LR$_{1;2;3}$} & 0.832 & 0.612/0.635/0.634 & 0.470/0.492/0.503 & 0.380/0.402/0.420 & 0.318/0.350/0.364 \\
		{\bf ICMH - Deep} & 0.943 & {\bf 0.768} & {\bf 0.672} & {\bf 0.601} & {\bf 0.542} \\ 
		{\bf ICMH - Deep + $\mathcal{L}_d$} & 0.943 & 0.761 & 0.653 & 0.587 & 0.537 \\ \hline \hline
		% *********************************************************************************
		CCCA$^*$ & 0.397 & 0.231 ($\overline{0.254}$, $\underline{0.226}$) & 0.154 ($\overline{0.181}$, $\underline{0.156}$) & 0.109 ($\overline{0.141}$, $\underline{0.118}$) & 0.091 ($\overline{0.115}$, $\underline{0.095}$) \\
		GMA$^*$ & 0.427 & 0.293 ($\overline{0.315}$, $\underline{0.264}$) & 0.207 ($\overline{0.239}$, $\underline{0.192}$) & 0.168 ($\overline{0.192}$, $\underline{0.152}$) & 0.138 ($\overline{0.160}$, $\underline{0.126}$) \\
		GSSL$^*$ & 0.672 & 0.362 ($\overline{0.518}$, $\underline{0.344}$) & 0.213 ($\overline{0.375}$, $\underline{0.223}$) & 0.158 ($\overline{0.293}$, $\underline{0.164}$) & 0.131 ($\overline{0.238}$, $\underline{0.131}$) \\
		SEPH$^*$ & 0.962 & 0.618 ($\overline{0.924}$, $\underline{0.602}$) & 0.474 ($\overline{0.858}$, $\underline{0.435}$) & 0.407 ($\overline{0.815}$, $\underline{0.343}$) & 0.341 ($\overline{0.776}$, $\underline{0.285}$) \\
		GSPH$_{klr}^*$ & 0.959 & 0.573 ($\overline{0.927}$, $\underline{0.582}$) & 0.425 ($\overline{0.865}$, $\underline{0.420}$) & 0.354 ($\overline{0.823}$, $\underline{0.323}$) & 0.301 ($\overline{0.781}$, $\underline{0.271}$) \\
		GSPH$_{lin}^*$ & 0.899 & 0.517 ($\overline{0.817}$, $\underline{0.514}$) & 0.340 ($\overline{0.726}$, $\underline{0.369}$) & 0.275 ($\overline{0.663}$, $\underline{0.197}$) & 0.230 ($\overline{0.606}$, $\underline{0.238}$) \\
		ACMR$^*$ & 0.932 & 0.291 ($\overline{0.935}$, $\underline{0.257}$) & 0.388 ($\overline{0.881}$, $\underline{0.179}$) & 0.333 ($\overline{0.839}$, $\underline{0.131}$) & 0.284 ($\overline{0.773}$, $\underline{0.104}$) \\
		GrowBit$^*$ & 0.955 & 0.465 ($\overline{0.937}$, $\underline{0.448}$) & 0.282 ($\overline{0.881}$, $\underline{0.279}$) & 0.231 ($\overline{0.817}$, $\underline{0.199}$) & 0.185 ($\overline{0.757}$, $\underline{0.152}$) \\
		%GrowBit$_{ce}^*$ & 0.963 & 0.546 ($\overline{0.949}$, $\underline{0.462}$) & 0.398 ($\overline{0.912}$, $\underline{0.290}$) & 0.338 ($\overline{0.887}$, $\underline{0.205}$) & 0.279 ($\overline{0.857}$, $\underline{0.159}$) \\
		SSCMLP$^*$ & 0.964 & 0.574 ($\overline{0.960}$, $\underline{0.323}$) & 0.506 ($\overline{0.927}$, $\underline{0.215}$) & 0.426 ($\overline{0.905}$, $\underline{0.157}$) & 0.375 ($\overline{0.886}$, $\underline{0.123}$) \\
		{\bf ICMH - LR$_{1;2;3}^*$} & 0.894 & 0.744/0.775/0.768 & 0.649/0.666/0.675 & 0.568/0.597/0.605 & 0.492/0.538/0.541 \\
		{\bf ICMH - Deep$^*$} & 0.962 & {\bf 0.825} & {\bf 0.754} & {\bf 0.694} & {\bf 0.658} \\ 
		{\bf ICMH - Deep + $\mathcal{L}_d^*$} & 0.959 & 0.817 & 0.737 & 0.686 & 0.650 \\ \hline
	\end{tabular}
	\label{rgbd_results}
\end{table*}

% **********************************************************************************

\section{Analysis of the algorithm}

Here we analyze the performance of our algorithm under different scenarios and also the effectiveness of the various adaptation strategies of the proposed framework.
% like - (1) length of hash codes (2) size of the exemplar set.
%We also try to understand how does each of the adaptation strategies are contributing to the performance improvement of our model under \textbf{P-III}.
Finally we provide the implementation details of our algorithm. \\ \\
% **********************************************************************************
{\bf Effect of varying the different hash codes:}
In all our experiments we have used the hash code size of $128$ bits to evaluate our proposed model.
Here we check the performance of our model when the number of hash codes are changed from $\{16,32,64,128\}$ bits. 
The results for cross-modal retrieval (average MAP@50 of the final stage) in Figure \ref{cm_bits} shows that with the increase in hash bits the performance significantly improves for all the four datasets in the incremental setting.
We observed the same improvement for the hashing also.
Interestingly using more number of hash bits seemed to give more improvements in the cross-modal experiments as compared to the hashing experiments.
This might be primarily due to the reason that in hashing we are using the non-regenerated hash codes for retrieval as opposed in the cross-modal retrieval where we have to generate the hash codes of the gallery set by passing it through the hash functions.\\ \\
% **********************************************************************************
{\bf Effect of varying the number of exemplars per category:}
Here, we evaluate the proposed framework for the two different datasets using hash bit of $128$ size and use different exemplar sizes of $\{5,10,20,30\}$ per class.
From the cross-modal retrieval results (average MAP@50 of the final stage) in Figure \ref{cm_exemplars}, we observe that as the number of exemplars increases, the performance in general increases at first and then saturates.
We observe from Figure \ref{cm_exemplars} that the performance of the three variants of ICMH-LR are lesser as compared to ICMH-Deep, especially for RGBD Object data.
As more number of exemplars per class of the old categories are available, the proposed framework performs better for both the old and new categories.
The same phenomenon has also been observed for the hashing experiments. \\ \\
% **********************************************************************************
{\bf Effect of imbalanced data sampler, class weighted losses for DNN:}
Here, we investigate how each strategy in our deep neural network strategy used in our updation scheme is helping our performance.
We perform the evaluation on the four datasets under four different conditions when we use - (1) only cross-entropy loss (2) cross-entropy loss + imbalanced data sampler (IBDS) (3) only weighted cross-entropy loss and (4)  weighted cross-entropy loss + imbalanced data sampler (IBDS).
We show the results in Table \ref{diff_loss_formula} (using $128$ bit hash code and $10$ exemplars per old categories) and observe that each component improves the performance whereas our final model gives the best result.
Interestingly enough we found that just simply using the weighted cross-entropy loss did not lead to a significant improvement but combining it with the imbalanced data sampler gave improvement.
This was found across all the four datasets and for both the cross-modal and hashing experiments.
We have also reported the results of ICMH - Deep when added with the distillation loss $\mathcal{L}^d$ \cite{icarl} \cite{lwf} \cite{lifelong} but did not find any significant improvements. \\ \\
\begin{table*}[t!]
	\tiny
	\centering
	\renewcommand{\arraystretch}{1.4}
	\setlength{\tabcolsep}{4.7 pt}
	\caption{MAP@50 and MAP@all for Wiki, LabelMe, Pascal Sentences and RGBD Object Database for the cross-modal retrieval and hashing experiments (denoted by ``*") under different updation strategies while training the deep neural network.
		We have trained the algorithms using $128$ hash code bit and an exemplar size of $10$ per old categories.
		We evaluate our deep neural network under four different conditions - (1) cross-entropy loss $\mathcal{L}_{ce}$, (2) weighted cross-entropy loss $\mathcal{L}_{wce}$, (3) $\mathcal{L}_{ce}$ + IBDS and (4) $\mathcal{L}_{ce}$ + IBDS.
		we observe that each component is helping to get better and better performance.}
	\label{diff_loss_formula}
	\begin{tabular}{|c|ccc|ccc|cccc|ccccc|}
		\hline
		\multicolumn{1}{|c|}{} & \multicolumn{15}{c|}{Dataset} \\ \hline
		Losses & \multicolumn{3}{c|}{Wiki} & \multicolumn{3}{c|}{LabelMe} & \multicolumn{4}{c|}{Pascal Sentences} & \multicolumn{5}{c|}{RGBD Object Database} \\ \hline
		Phases & 1 & 2 & 3 & 1 & 2 & 3 & 1 & 2 & 3 & 4 & 1 & 2 & 3 & 4 & 5 \\ \hline
		$\mathcal{L}_{ce}$ & 0.655 & 0.437 & 0.327 & 0.906 & 0.695 & 0.543 & 0.677 & 0.566 & 0.466 & 0.387 & 0.941 & 0.632 & 0.488 & 0.432 & 0.356 \\ 
		$\mathcal{L}_{wce}$ & 0.660 & 0.435 & 0.302 & 0.901 & 0.679 & 0.562 & 0.670 & 0.575 & 0.455 & 0.393 & 0.939 & 0.620 & 0.474 & 0.435 & 0.358 \\ 
		$\mathcal{L}_{ce}$ + IBDS & 0.647 & 0.475 & 0.380 & 0.898 & 0.778 & 0.686 & 0.672 & 0.592 & 0.512 & 0.431 & 0.946 & 0.760 & 0.666 & 0.594 & 0.547 \\
		$\mathcal{L}_{wce}$ + IBDS & 0.659 & 0.480 & 0.384 & 0.906 & 0.767 & 0.694 & 0.670 & 0.589 & 0.504 & 0.427 & 0.943 & 0.768 & 0.672 & 0.601 & 0.542 \\
		$\mathcal{L}_{wce}$ + IBDS + $\mathcal{L}_{d}$ & 0.655 & 0.479 & 0.390 & 0.906 & 0.781 & 0.710 & 0.670 & 0.587 & 0.496 & 0.420 & 0.943 & 0.761 & 0.653 & 0.587 & 0.537 \\ \hline \hline
		$\mathcal{L}_{ce}^*$ & 0.841 & 0.640 & 0.539 & 0.950 & 0.787 & 0.699 & 0.814 & 0.722 & 0.660 & 0.586 & 0.961 & 0.678 & 0.551 & 0.505 & 0.426 \\ 
		$\mathcal{L}_{wce}^*$ & 0.842 & 0.635 & 0.529 & 0.951 & 0.789 & 0.708 & 0.805 & 0.718 & 0.645 & 0.585 & 0.962 & 0.679 & 0.546 & 0.511 & 0.431 \\ 
		$\mathcal{L}_{ce}$ + IBDS$^*$ & 0.838 & 0.664 & 0.579 & 0.951 & 0.850 & 0.784 & 0.810 & 0.749 & 0.696 & 0.637 & 0.960 & 0.817 & 0.741 & 0.688 & 0.655 \\
		$\mathcal{L}_{wce}$ + IBDS$^*$ & 0.837 & 0.674 & 0.595 & 0.951 & 0.853 & 0.795 & 0.807 & 0.745 & 0.693 & 0.640 & 0.962 & 0.825 & 0.754 & 0.694 & 0.658 \\
		$\mathcal{L}_{wce}$ + IBDS + $\mathcal{L}_{d}^*$ & 0.837 & 0.675 & 0.588 & 0.952 & 0.858 & 0.801 & 0.807 & 0.749 & 0.688 & 0.629 & 0.959 & 0.817 & 0.737 & 0.686 & 0.650 \\ \hline
	\end{tabular}
\end{table*}
% **********************************************************************************
{\bf Does this adaptation strategies help other baseline algorithms in the incremental setting ?}
We try to understand whether our strategies can help in the incremental setting for other baseline algorithms here.
Interestingly enough SSCMLP \cite{sscmlp} cannot be modified to work in an incremental fashion as the embedding/common space in the algorithm has the same dimension as the number of classes. 
Since in this protocol, with the arrival of subsequent training data, the number of categories increases, the common dimension would be needed to be expanded itself which makes adapting the model to new categories while retaining the old category information difficult.
In converse, since the ACMR \cite{acmr} algorithm has a common domain representation whose dimension is independent of the number of classes, our adaptation strategies can be easily included in its formulation.

We have considered the basic ACMR \cite{acmr} algorithm and added the adaptation strategies like weighted losses, imbalanced datasampler and distillation loss \cite{icarl} \cite{lifelong} \cite{lwf} to the algorithm.
We report the best results of the modified ACMR \cite{acmr} which we obtained by using this tricks in Table \ref{acmr_incremental} on the Wiki \cite{wiki} and LabelMe \cite{labelme} datasets.
We have observed here also that there is a significant improvement in the results for both the cross-modal retrieval and hashing experiments.
We notice that now the performance gap between the modified ACMR \cite{acmr} algorithm has greatly reduced as compared to our ICMH algorithm for the cross-modal experiments though in the hashing domain ICMH performs significantly better than the modified ACMR algorithm.
This is primarily due to the fact for the hashing paradigm we were able to use the non-regenerated hash codes for the ICMH algorithm. \\ \\
\begin{table}[t!]
	\tiny
	\centering
	\renewcommand{\arraystretch}{1.4}
	\setlength{\tabcolsep}{18.0 pt}
	\caption{MAP@50 and MAP@all for Wiki and LabelMe dataset for the cross-modal retrieval and hashing experiments (denoted by ``*") using different adaptation strategies for the ACMR \cite{acmr} algorithm.
		We report the best results for the modified ACMR algorithm here.}
	\label{acmr_incremental}
	\begin{tabular}{|c|ccc|ccc|}
		\hline
		\multicolumn{1}{|c|}{} & \multicolumn{6}{c|}{Dataset} \\ \hline
		Losses & \multicolumn{3}{c|}{Wiki} & \multicolumn{3}{c|}{LabelMe} \\ \hline
		Phases & 1 & 2 & 3 & 1 & 2 & 3 \\ \hline
		ACMR & 0.718 & 0.432 & 0.314 & 0.932 & 0.724 & 0.595 \\
		modified ACMR  & 0.710 & 0.487 & 0.372 & 0.931 & 0.807 & 0.683 \\
		{\bf ICMH -  Deep} & 0.659 & 0.480 & 0.384 & 0.906 & 0.767 & 0.694 \\
		{\bf ICMH -  Deep} + $\mathcal{L}_{d}$ & 0.655 & 0.479 & 0.390 & 0.906 & 0.781 & 0.710 \\ \hline \hline
		ACMR$^*$ & 0.823 & 0.493 & 0.297 & 0.949 & 0.648 & 0.521 \\ 		
		modified ACMR$^*$  & 0.830 & 0.611 & 0.452 & 0.950 & 0.763 & 0.608 \\ 
		{\bf ICMH -  Deep}$^*$ & 0.837 & 0.674 & 0.595 & 0.906 & 0.767 & 0.694 \\
		{\bf ICMH -  Deep} + $\mathcal{L}_{d}^*$ & 0.837 & 0.675 & 0.588 & 0.906 & 0.781 & 0.710 \\	\hline
	\end{tabular}
\end{table}
% **********************************************************************************
{\bf Implementation details:}
We have implemented the non-deep and deep algorithm in Matlab and PyTorch \cite{pytorch} respectively.
%For proper implementation in Matlab we build upon the prior works in \cite{gsph} and our deep model adaptation strategy is based on the preliminary works in \cite{growbit}.
For training ICMH-LR, the hyper-paramters $\lambda_{lr}^x, \lambda_{lr}^y, \gamma_x, \gamma_y$ are set using five-fold cross-validation.
For ICMH-Deep, the model is trained using SGD with a learning rate of $\{ \{0.001,0.01\}, \{0.1,0.1\}, \{0.1,0.1\}, \{0.001,0.001\} \}$ for $\{ 500, 200, 500, 1000 \}$ epochs on the Wiki, LabelMe, Pascal Sentences and RGBD Object dataset (two domains) respectively.
The networks details are provided in Section 4.2.2 with hidden layer of size $\{ \{500,250\}, \{500,250\}, \{500,250\}, \{1000,1000\} \}$ for $\{ 500, 200, 500, 1000 \}$ for the four datasets respectively.
The weight parameter $w_j$ has been set by studying the frequency of occurrence of each class in the training data.

% **********************************************************************************
\section{Conclusion}

In this work we analyze how cross-modal algorithms behave in retrieval and hashing applications when it is exposed to new categories of data.
We establish three protocols to analyze the performance issues related to this and explore strategies regarding how to update the model when new categories of data are suddenly made available.
Here, we have proposed both non-deep and deep based approaches and experimented with knowledge based distillation losses, weighted classification loss and imbalanced data sampling to mitigate the effects of catastrophic forgetting by using the information contained in the old categories through the storage of exemplars.
We have performed extensive experiments using a variety of baseline cross-modal algorithms on four standard datasets to understand the difficulty of trained model to adapt to the data over the new categories.
Experiments using our proposed model shows that our approach can somewhat mitigate the effects of catastrophic forgetting when exposed to new categories of data.

\bibliography{egbib}
\end{document}